\documentclass{article}

\usepackage{arxiv}
\renewcommand{\undertitle}{}

\usepackage[utf8]{inputenc} 
\usepackage[T1]{fontenc}    
\usepackage{hyperref}       
\usepackage{url}            
\usepackage{booktabs}       
\usepackage{amsfonts}       
\usepackage{nicefrac}       
\usepackage{microtype}      
\usepackage{lipsum}		
\usepackage{graphicx}
\usepackage{float}
\usepackage[numbers,square]{natbib}
\usepackage{doi}

\title{Nested Shared-Operator Recursion on ARC-AGI-1: 
An Empirical Study of HRM, TRM and URM}

\author{
Sara Neri \\
\normalfont\small University of Padova \\
\normalfont\small Padova, Italy \\
\normalfont\small Universidad Politécnica de Madrid \\
\normalfont\small Madrid, Spain \\
\normalfont\footnotesize\texttt{sara.neri01@gmail.com}
\And
Eugenio Gaeta \\
\normalfont\small Escuela Técnica Superior de \\
\normalfont\small Ingenieros de Telecomunicación \\
\normalfont\small Universidad Politécnica de Madrid \\
\normalfont\small Av. Complutense, 30 \\
\normalfont\small 28040 Madrid, Spain \\
\normalfont\footnotesize\texttt{eugenio.gaeta@upm.es}
\And
Andrea Ceschini \\
\normalfont\small University of Rome ``La Sapienza'' \\
\normalfont\small Via Eudossiana 18 \\
\normalfont\small 00184 Rome, Italy \\
\normalfont\footnotesize\texttt{andrea.ceschini@uniroma1.it}
\AND
Andrea Facchinetti \\
\normalfont\small Department of Information Engineering (DEI) \\
\normalfont\small University of Padova \\
\normalfont\small Via G. Gradenigo 6/B \\
\normalfont\small 35131 Padova, Italy \\
\normalfont\footnotesize\texttt{andrea.facchinetti@unipd.it}
\And
Giuseppe Fico \\
\normalfont\small Escuela Técnica Superior de \\
\normalfont\small Ingenieros de Telecomunicación \\
\normalfont\small Universidad Politécnica de Madrid \\
\normalfont\small Av. Complutense, 30 \\
\normalfont\small 28040 Madrid, Spain \\
\normalfont\footnotesize\texttt{gfico@lst.tfo.upm.es}
}

\begin{document}
\maketitle
\begin{abstract}
	Recursive reasoning models refine persistent internal states by repeatedly applying learned operators. We investigate whether adding a persistent intermediate state that integrates repeated updates of a faster state through the same operator improves predictive performance relative to each model family’s baseline at matched parameter counts. We define this organisation operationally and implement it within three architectures: the Hierarchical Reasoning Model (HRM), Tiny Recursive Model (TRM) and Universal Reasoning Model (URM). Using the official codebases on ARC-AGI-1, we compare each baseline with its parameter-matched nested variant. We evaluate held-out queries from tasks whose demonstrations were available during training. The experiments use a single-GPU budget far below the published training protocols. Under the tested conditions, nested variants show no consistent advantage in token accuracy, predictive loss or exact solving, which remains near zero for all configurations. The results therefore leave the hypothesis open. We specify a confirmatory comparison, with an informative baseline, multiple seeds, controls that isolate nesting and predefined criteria, to determine whether this organisation merits further development.

\end{abstract}


\section{Introduction}
Recent progress in reasoning with neural networks has renewed interest in how much computation a model performs before producing an answer and in how that computation is organised. In language models, additional computation is often obtained through longer autoregressive sequences in which intermediate steps are represented as generated tokens \cite{wei_chain--thought_2022}. A different line of work moves part of the inference into latent space \cite{hao_training_2024,geiping_scaling_2025}. These approaches iteratively update internal representations before producing an output. Latent iteration can increase effective computational depth without lengthening the linguistic reasoning trace. The organisation of iterative computation thus becomes an architectural variable in its own right.

Recursive state updating is one concrete form of latent iterative computation. A learned transformation is repeatedly applied to an internal state that persists across computational steps. Reusing the same parameters can increase effective computational depth without a proportional increase in parameter count \cite{lan_albert_2020,dehghani_universal_2019}. Parameter sharing may also introduce an inductive bias by constraining the model to reuse a common transformation across successive updates. These properties depend on the training procedure. Training deeply unrolled recurrent computations can be difficult because gradients may vanish or explode \cite{pascanu_difficulty_2013,bengio_learning_1994}.

Three architectures illustrate different organisations of recursive computation. The Hierarchical Reasoning Model (HRM) \cite{wang_hierarchical_2025} uses two interdependent recurrent modules operating at different update frequencies. The Tiny Recursive Model (TRM) \cite{jolicoeur-martineau_less_2025} uses a single shared network to refine a latent state and a candidate solution. The Universal Reasoning Model (URM) \cite{gao_universal_2025} builds on a shared Universal Transformer transition augmented with short convolution. All three report strong performance on structured reasoning benchmarks, but differ in architecture, gradient propagation and training procedures. 

Existing evidence does not isolate the effect of recursive organisation. HRM, TRM, and URM differ simultaneously in network structure, parameter count, gradient propagation, halting, stabilisation, data augmentation, and inference procedure. Reported benchmark results reflect complete architectural and training configurations rather than a single controlled factor. This makes it difficult to determine whether performance differences arise from the organisation of recursive computation or from other design choices that vary across models.

We investigate this question through Fractal Recursion, an organisation that nests recursive updates over persistent states using shared operators. 
The motivating intuition is that nested updates may support refinement at different computational scales. A persistent intermediate state can integrate successive updates of a faster state and, in turn, condition its subsequent refinement. This organisation could make parameter reuse more effective by allowing information to be maintained and updated at different frequencies. Whether these states learn useful, complementary functions remains an empirical question. 

The hypothesis examined in this study is that adding a persistent intermediate state and organising lower-level recursive updates around it, using the same operator for the fast and intermediate states, improves predictive performance relative to each model family’s baseline at matched parameter counts. A controlled test of this hypothesis requires matched parameter counts and comparable data and computational budgets. The present comparisons preserve parameter counts within each family, but do not fully match computational budgets. The present experiments evaluate specific implementations of this idea, with family-specific differences in computation, input coupling and gradient propagation that limit attribution of the observed outcomes to nesting alone.

The main comparisons use the first version of the Abstraction and Reasoning Corpus for Artificial General Intelligence (ARC-AGI-1) \cite{chollet_measure_2019,chollet_arc_2025} and evaluate held-out queries from tasks whose demonstrations were available during training. They therefore assess prediction of unseen query outputs, rather than generalisation to entirely unseen tasks. 

The study is conducted under a resource-constrained computational regime relative to the original model-development efforts, and its evidence is interpreted as a constrained test of the architectural hypothesis rather than as a full-scale reproduction. 
This work is motivated by a conceptual framework under development within our research group, termed Reasoning Foundation Models (RFMs), in which reasoning proceeds through recursive updates of internal states over a structured problem representation. 

The framework distinguishes a representation that organises problem information from an internal Reasoning State updated by a recursive core, and is broader than the architectures evaluated here. HRM, TRM and URM provide concrete mechanisms for repeated latent-state updating, but they operate on inputs already given in structured form and do not implement the complete framework; their internal states are not assumed to correspond to semantically grounded Reasoning States. We use them as settings in which to examine recursive-core organisation. Fractal Recursion is one hypothesis about the recursive core and is not a requirement of the RFM framework.

This work makes four contributions. First, it provides an operational definition of Fractal Recursion, distinguishing it from longer recurrence and from replicated modules with independent weights, and specifies the nested shared-operator configurations examined here. Second, it implements parameter-matched baseline–variant comparisons within HRM, TRM and URM, since published comparisons across these models involve multiple architectural and training differences. Third, it documents update schedules, operator-application counts, gradient propagation and remaining confounding factors, showing that parameter matching alone does not isolate nesting. Fourth, it reports no consistent predictive advantage on ARC-AGI-1 under the tested conditions and outlines a controlled experimental test with explicit criteria for supporting, weakening or leaving the hypothesis unresolved.

\section{Related Work}
\label{sec:headings}

Several lines of work have explored how computational depth can be increased through repeated application of learned transformations. The Universal Transformer reuses a recurrent self-attentive transition across depth and combines it with adaptive halting \cite{dehghani_universal_2019}. Deep Equilibrium Models represent an extreme form of weight tying, defining the output through the fixed point of a shared nonlinear transformation \cite{bai_deep_2019}. Looped Transformers similarly reuse the same block to support iterative computation \cite{giannou_looped_2023,yang_looped_2024,saunshi_reasoning_2025}. More recently, recurrent-depth language models have extended this principle to latent reasoning by iterating a recurrent block and scaling computation at test time \cite{geiping_scaling_2025}. These approaches establish parameter reuse across computational depth. Fractal Recursion considers an additional variable, namely how repeated updates are organised across nested computational scales.

Independent analysis of HRM further motivates separating hierarchical structure from recursive organisation. The ARC Prize Team \cite{arc_prize_team_hidden_2025} replaced HRM with a Transformer of the same parameter count, approximately 27M, while keeping the remaining training pipeline unchanged. The Transformer came within about five percentage points of HRM without additional hyperparameter optimisation, while the outer refinement loop produced substantially larger performance gains. The comparison was parameter-matched but not compute-matched, since HRM used more computation. These findings suggest that the specific high-level and low-level hierarchy should not be assumed to account for HRM's performance by itself. HRM also relies on two recurrent modules with distinct parameters. The analysis therefore does not test whether a shared recursive operator benefits from being organised across nested self-similar scales.

Published scores for HRM, TRM and URM reflect different training, augmentation, halting, inference and evaluation protocols. They therefore do not isolate recursive organisation as an architectural variable; this requires within-family comparisons with explicit controls.
TRM also reports a multi-scale variant that replaces its single latent feature with several features, each updated once per recursion and conditioned on all others; on Sudoku-Extreme this variant performed worse than the two-feature design (77.6\% versus 87.4\%) \cite{jolicoeur-martineau_less_2025}. In that variant, no slower state integrates repeated cycles of a faster one. The variants studied here instead nest repeated fast-state updates within each intermediate-state update, using the same operator.

Clockwork RNN \cite{koutnik_clockwork_2014} uses recurrent modules with predefined update rates, while Hierarchical Multiscale RNN \cite{chung_hierarchical_2017} learns temporal scales through differently updated recurrent layers. These approaches motivate distinguishing update frequencies from computational depth. Here, we examine nested internal updates over a fixed problem representation, with shared or parameter-tied operators across scales.

FractalNet \cite{larsson_fractalnet_2017} constructs self-similar feed-forward architectures through a recursive expansion rule, while Fractal Generative Models \cite{li_fractal_2025} recursively compose generative modules. These studies concern architectural or generative self-similarity; our intervention concerns the organisation of updates over persistent computational states.

Adaptive Computation Time (ACT) \cite{graves_adaptive_2016} learns when recurrent processing should halt, while PonderNet \cite{banino_pondernet_2021} learns a distribution over processing depth. These mechanisms regulate computational quantity and can complement the study of how updates are organised across nested scales.
Prior work provides shared transformations across depth, recurrent computation at multiple timescales, self-similar architectures and adaptive computational depth. Fractal Recursion combines a specific subset of these properties, defined operationally in the Methodology section.

\section{Methodology}
\label{sec:others}


In this study, Fractal Recursion is defined by four architectural conditions: computational self-similarity, nesting, operator sharing and state persistence. Parameter parity is an additional experimental constraint on the variants evaluated.

The term fractal refers here to an architectural analogy: a shared update operator is reused within a nested organisation of persistent states at distinct computational scales. It does not imply a demonstrated mathematical property of self-similarity or scale invariance. The four conditions below specify the organisation considered in this study; they are not, by themselves, a claim of novelty relative to all existing multiscale recurrent architectures.\\
\textbf{Computational self-similarity}. The same recursive update principle is reproduced at more than one computational scale.\\
\textbf{Nesting}. At least two distinguishable recursive scales are organised within one another, rather than forming a single longer sequence of updates.\\
\textbf{Operator sharing}. Corresponding updates across scales reuse the same learned operator or parameter-tied operators.\\
\textbf{State persistence}. Internal states persist across recursive updates and transmit information between connected scales.\\
\textbf{Experimental constraint — parameter parity}. Introducing the nested organisation does not increase the number of trainable parameters relative to the within-family reference configuration.

The evaluated variants are designed to satisfy these four conditions under a parameter-parity constraint. We denote the nested variants by HRM-F, TRM-F and URM-F, where the suffix -F refers to Fractal Recursion. The intervention concerns the organisation of lower-level updates within each model family. HRM and TRM already contain outer and inner update schedules. In particular, the TRM baseline satisfies the four architectural conditions above, since repeated updates of L are nested within cycles that update H, with both persistent states updated through the same operator. The conditions therefore characterise the organisation considered here, rather than uniquely distinguishing the proposed variants from their baselines. The proposed variants introduce an additional persistent intermediate state and reorganise lower-level updates into repeated groups in which fast-state updates precede an intermediate-state update using the same lower-level operator or shared core. Increasing iteration counts alone does not introduce this organisation, while replicated modules with independent weights do not satisfy the operator-sharing condition.
Each nested variant is compared with its within-family baseline. Parameter counts, operator-application counts and differentiable update segments are documented separately. These checks establish parameter parity and describe computational and gradient-path differences; they do not establish equal runtime, memory use or overall computational cost.
For the official-codebase comparisons, e denotes the embedded input, including puzzle embeddings. Each operator denotes a complete layer stack; additions are elementwise and assignments are sequential, using the latest available states.\\

\textbf{HRM.} The baseline repeats $L \leftarrow F_L(L + H + e)$ twice before updating $H \leftarrow F_H(H + L)$. The nested variants replace each lower update with two groups of
\[
L \leftarrow F_L(L + M + H + e), \quad
L \leftarrow F_L(L + M + H + e), \quad
M \leftarrow F_L(M + L),
\]
and subsequently update $H \leftarrow F_H(H + M)$. In the variants, $H$ is therefore updated from $M$ rather than from $L$. The lower stack is shared between $L$ and $M$; the upper stack remains separate. Each $H$ cycle therefore contains two lower-stack applications in the baseline and twelve in the variants (\textit{Figure 1a}). With two $H$ cycles per segment, the totals are four and twenty-four lower-stack applications, respectively, plus two upper-stack applications. The baseline differentiates only the final $L$ and $H$ updates. C1 differentiates the final $L$, $M$ and $H$ updates; C2 differentiates all six lower-stack applications in the final nested macro-update and the final $H$ update. Earlier updates run without gradient tracking. C1 and C2 have the same forward schedule.\\

\textbf{TRM.} The baseline executes six updates $L \leftarrow F(L + H + e)$, followed by $H \leftarrow F(H + L)$. The variant instead executes two groups of
\[
L \leftarrow F(L + M + e), \quad
L \leftarrow F(L + M + e), \quad
M \leftarrow F(M + L),
\]
followed by $H \leftarrow F(H + M)$. All updates share one stack. Three macro-cycles give twenty-one applications per segment in both configurations; the first two cycles run without gradient tracking and all seven applications in the final cycle are differentiable. Besides adding an intermediate state, the variant removes $H$ from the lower-state updates and reduces direct input injection from six to four applications per macro-cycle. Thus, matched application counts do not isolate nesting from changes in state coupling.\\

\textbf{URM.} The baseline repeatedly updates a single state through $z \leftarrow F(z + e)$. Each of its two groups contains six applications. The variant replaces each group with two $L$--$L$--$M$ sequences using the TRM variant's lower-state equations above, without an $H$ state. Both configurations therefore execute twelve shared-stack applications per segment. The first group runs without gradient tracking; all six applications in the second are differentiable. Predictions use $z$ in the baseline and $M$ in the variant. This examined configuration differs from the eight inner-loop steps, the first two forward-only, reported for the original URM. Direct input reaches every baseline update but only the fast-state updates in the variant.

In official HRM and TRM, H and L are initialised from fixed, non-trainable buffers; the added M state uses the same initial value as L. Official URM similarly initialises both variant states from its baseline’s fixed initial-state buffer. States retain their values across internal cycles and successive ACT segments, but the returned carry is detached, preventing gradients from crossing segment boundaries. The wrapper resets states and replaces data separately for each previously halted batch row. HRM and TRM prediction and halting heads read H; URM heads read its output state defined above. Initial-state buffers introduce no additional trainable parameters.

These schedules count applications along the main state trajectory. Official HRM training additionally executes a complete internal segment without gradient tracking to construct the continuation target. Overall cost therefore also depends on auxiliary computation and the number of ACT segments executed.
Reported parameter counts refer to the network and exclude the separately learned puzzle-embedding tables. In the examined official configurations, HRM has 27,276,290 network parameters and 448,720,384 puzzle-embedding parameters (475,996,674 combined); TRM has 6,829,058 and 448,719,872, respectively (455,548,930 combined); URM has 13,663,234 and 448,720,384 (462,383,618 combined). Counts match between baseline and nested variants within each family, both for the network and for the puzzle-embedding table. Fixed initial-state buffers are excluded. The examined adapted HRM and TRM configurations have 7,535,712 and 9,507,936 parameters, respectively, with matched counts within each pair and no learned puzzle-embedding tables.

The computational controls differ across families. HRM retains parameter matching while increasing lower-operator applications. TRM and URM retain the number of core applications. These counts do not establish identical runtime or memory use. Total computation additionally depends on the steps executed by the ACT wrapper.

\begin{figure}[H]
    \centering
    \includegraphics[width=\textwidth]{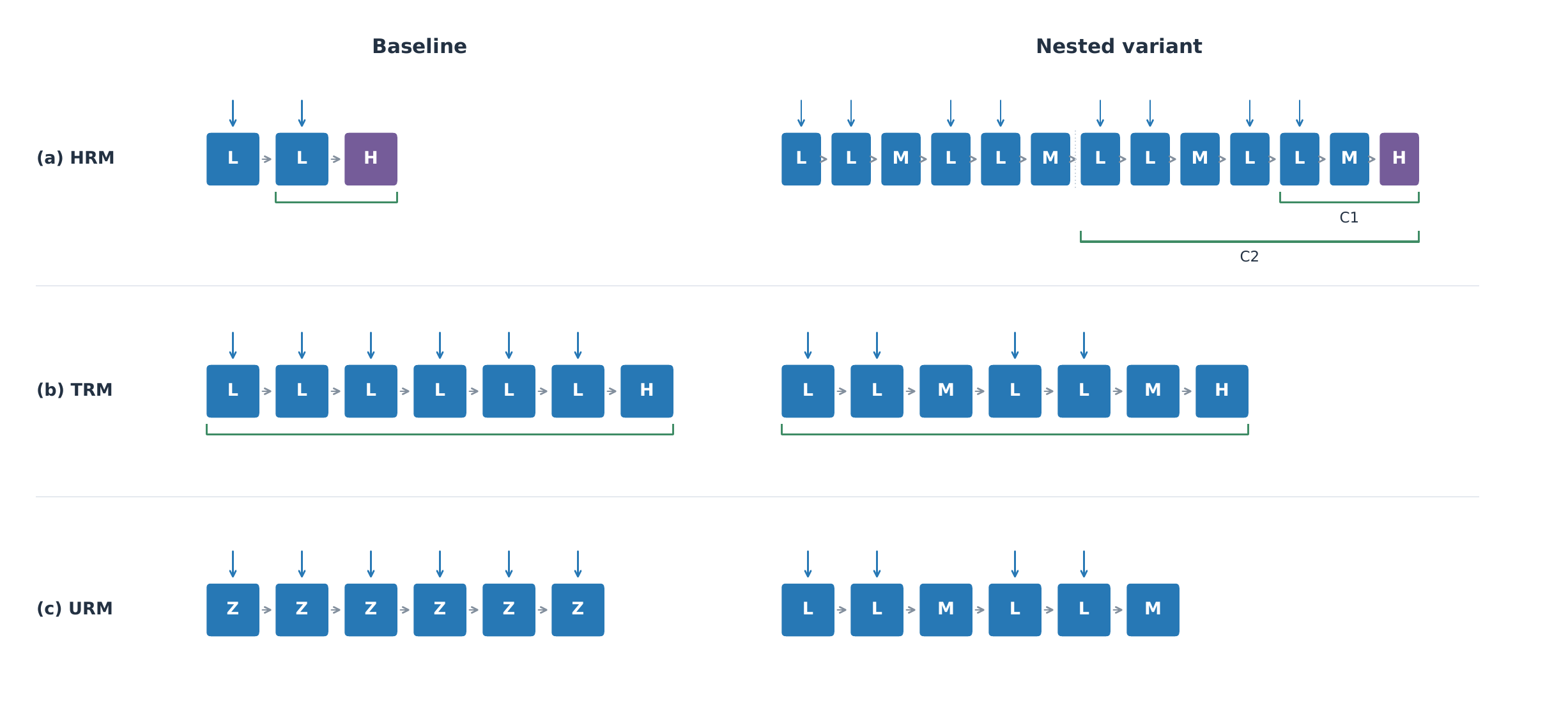}
    \caption{\textit{Update schedules for baseline and nested configurations. Boxes denote state updates; horizontal arrows indicate execution order, not all connections between states. Blue boxes within each family share an operator; purple denotes HRM’s separate upper operator. Downward arrows mark direct input injection. \textbf{(a) HRM}: one H cycle contains two lower-operator applications in the baseline and twelve in the variant, organised as two lower macro-steps of six. Green brackets show differentiable applications in the final H cycle: the last L followed by H in the baseline; the last two lower-operator applications (L and M), followed by H, in C1; and all six applications of the final lower macro-step, followed by H, in C2. Input enters L updates directly, but not M or H. \textbf{(b) TRM}: seven shared-core applications per macro-cycle, repeated over three cycles, give 21 applications per segment. Direct input enters six updates in the baseline and four in the variant. Green brackets refer to the final cycle; the preceding two run without gradient tracking. H denotes the candidate-solution update. \textbf{(c) URM}: one group of six core applications is shown for each configuration. Each ACT segment executes two such groups in sequence; the first runs without gradient tracking, and all six applications in the second are differentiable.
}}
    \label{fig:figure1}
\end{figure}

Preliminary development used adapted HRM and TRM implementations, termed rapid, with explicit demonstration conditioning. They supported implementation checks and exploratory optimisation and are described, with their results, in Appendix A.

The main comparisons evaluate performance on ARC-AGI-1 using the preprocessed datasets specified for each official implementation. Before augmentation, the HRM and TRM training sets contain 960 task groups: the 400 ARC-AGI-1 training tasks, the demonstrations of the 400 evaluation tasks and 160 ConceptARC tasks \cite{moskvichev_conceptarc_2023}, for a total of 3,988 original training pairs. The evaluation set contains the 419 test queries of the 400 evaluation tasks. URM uses the same training files as HRM; the TRM dataset build contains one fewer puzzle identifier and 16 fewer augmented training sequences (3,633,410 versus 3,633,426). Demonstrations from evaluation tasks are included in training, while their test queries are reserved for evaluation \cite{arc_prize_team_hidden_2025}. The experiments therefore assess held-out query outputs from tasks whose demonstrations were available during training, rather than generalisation to entirely unseen tasks. Availability of task demonstrations during training is part of this evaluation regime and must be distinguished from access to the held-out query targets. In the official implementations, each task is associated with learned puzzle embeddings that are combined with the input; the demonstrations of evaluation tasks provide the training signal for the embeddings later used with their held-out queries. The adapted implementations instead condition explicitly on demonstrations and use no learned puzzle embeddings.
Grids are encoded on a 30 × 30 canvas, using token 0 for padding, token 1 for end-of-sequence markers, and tokens 2–11 for ARC colours 0–9. Augmentation combines the eight dihedral transformations with colour permutations that leave ARC colour 0 unchanged. Training preprocessing additionally applies joint translations to input–output pairs; test sequences are encoded without translation. The preprocessing requests up to 1,000 additional augmented instances per original task, but deduplication yields fewer instances for some tasks.

Training uses AdamATan2 \cite{everett_scaling_2024} for model parameters and a separate SignSGD optimiser \cite{bernstein_signsgd_2018} for puzzle embeddings. The learning rate is 10$^{-4}$ for model parameters and 10$^{-2}$ for puzzle embeddings, with 2,000 warm-up steps followed by constant learning rates. Optimisation settings are shared within each pair. Weight decay is 0.1 for both model parameters and puzzle embeddings in all official-codebase runs. HRM and TRM are evaluated using ordinary checkpoint weights, without exponential moving averaging (EMA). URM checkpoints contain EMA model parameters with a decay coefficient of 0.999; puzzle embeddings are not averaged. The TRM runs used a global batch size of 32 and no EMA.\\

In the examined official implementations, the predictive objective uses stablemax cross-entropy \cite{prieto_grokking_2025}, averaged over non-ignored target positions within each sequence and then summed over the batch. Target tokens with ID 0 are mapped to the ignore index. HRM adds $0.5(\mathcal{L}_{\mathrm{halt}} + \mathcal{L}_{\mathrm{continue}})$, whereas the examined TRM and URM configurations add only $0.5\mathcal{L}_{\mathrm{halt}}$. These auxiliary terms supervise the halting and, where active, continuation predictions. The resulting objective is divided by the global batch size before backpropagation; URM additionally accounts for gradient accumulation, whose factor is one in the examined runs.\\

\begin{table}[H]
\centering
\small
\setlength{\tabcolsep}{4pt}
\renewcommand{\arraystretch}{1.3}
\begin{tabular}{|p{0.20\textwidth}|p{0.17\textwidth}|p{0.20\textwidth}|p{0.14\textwidth}|p{0.19\textwidth}|}
\hline
\textbf{Configurations} &
\textbf{Global batch size} &
\textbf{Training budget} &
\textbf{Available seeds} &
\textbf{Optimiser updates} \\
\hline
HRM baseline and HRM-F C2 &
48 &
61 epochs &
0, 1 &
4,886 (seed 0); 4,890 (seed 1) \\
\hline
HRM-F C1 &
48 &
61 epochs &
0 &
4,886 \\
\hline
TRM baseline and TRM-F &
32 &
50,000 optimisation steps &
0 &
50,000 \\
\hline
URM baseline and URM-F &
48 &
60 epochs &
0 &
4,807 \\
\hline
URM baseline and URM-F, extended budget &
48 &
120 epochs, trained from scratch &
0 &
9,614 \\
\hline
\end{tabular} 

\vspace{8pt}

\caption{\textit{Training configurations and budgets for the official-codebase comparisons. For HRM and URM, an epoch denotes one shuffled traversal of the dataset groups, with one puzzle sampled per group and examples drawn from that puzzle to fill batches, rather than a full pass over all stored augmented sequences. Incomplete final batches are discarded at the end of each loader iteration, which may span multiple epochs. Optimiser updates correspond to the steps of the evaluated checkpoints.}}
\label{tab:training}
\end{table}

The study was conducted under a single-GPU research budget, with limited time for training and repeated runs. The available infrastructure included an H100 Multi-Instance GPU (MIG) partition with approximately 47 GB of GPU memory. For context, the published TRM ARC protocol reports 100,000 training epochs, a batch size of 768 and approximately three days on four 80 GB H100 GPUs. For reference, the original publications report ARC-AGI-1 scores of 40.3\% for HRM and 44.6\% for TRM under the two-attempt metric, and 53.8\% pass@1 for URM, each under its own training and evaluation protocol; the present baselines are not reproductions of those results. Our adopted budgets are reported in Table 1; epochs and optimisation steps are not directly equivalent across codebases. These resource constraints delimit the regimes and repetitions examined, but do not establish that additional training would favour the nested variants.\\
Evaluation distinguishes partial token correctness, exact augmented-sequence prediction and exact solving of original queries or tasks. In the official pipelines, mean per-sequence token accuracy is calculated by computing the fraction of correctly predicted tokens at valid labelled positions within each sequence and then averaging over valid examples. Valid positions may include delimiter tokens. We use this name for the official accuracy results throughout the paper. Augmented-sequence exact accuracy requires every valid labelled position in an augmented sequence to be correct and is averaged over evaluated augmented sequences. Neither measure directly gives the fraction of original tasks solved. Original-query and original-task exact-solving measures use the distinct aggregation rules specified in Table~2.

Original-query evaluation reconstructs the largest-area rectangle anchored at the top-left corner that contains only colour tokens, then reverses the geometric transformation and colour permutation. This reconstruction does not explicitly reject malformed predictions: it may return an empty grid or recover a rectangular crop. Predictions are grouped by task and original query. Identical reconstructed grids are deduplicated, retaining their frequencies, and ranked by decreasing frequency,  with mean halting confidence $\sigma(q_h)$ used to break ties. Here, $k$ denotes the number of distinct ranked candidates considered, rather than the number of augmentations or ACT segments. Scores at $k = 1000$ measure whether the correct output is recovered within a large candidate list, rather than whether the model reliably selects it as its first answer. A query is solved at $k$ when its expected output appears among those candidates; aggregation across queries and tasks follows Table~2.

The exact-correctness criteria and aggregation units are summarised in the following table and kept distinct throughout the analysis.\\

\begin{table}[H]
\centering
\small
\setlength{\tabcolsep}{4pt}
\renewcommand{\arraystretch}{1.3}
\begin{tabular}{|p{0.23\textwidth}|p{0.31\textwidth}|p{0.39\textwidth}|}
\hline
\textbf{Evaluation} &
\textbf{Exact-correctness criterion} &
\textbf{Aggregation} \\
\hline
Augmented-sequence exact accuracy &
All valid labelled positions in a sequence are correct. &
Mean over evaluated augmented sequences. \\
\hline
HRM original-task score at $k$ &
Every test query of a task has its expected output among its first $k$ selected candidates. &
Fraction of the 400 tasks entirely solved; solved queries are also counted globally over the 419 test queries. Assessed for baseline and C2 at seed 0. \\
\hline
TRM/URM ARC/pass@$k$ &
A query's expected output is among its first $k$ selected candidates. &
Fraction of solved queries within each task, averaged equally across the 400 tasks. \\
\hline
\end{tabular}
\vspace{8pt}

\caption{\textit{Exact-correctness criteria and aggregation units. These measures are not interchangeable: a global count of solved queries is not equivalent to the per-task mean used by ARC/pass@$k$, and neither equals the fraction of entirely solved tasks. $k$ denotes the number of distinct ranked candidates; first-candidate scores correspond to $k = 1$.}}
\label{tab:evaluation}
\end{table}

URM additionally reports a metric named ARC/maj@$N$ in the code. This procedure samples $N$ predictions with replacement, with probabilities proportional to halting confidence, and selects the sampled prediction with the highest halting confidence. It does not implement majority voting. Exact correctness is averaged first across queries within each task and then equally across tasks. Here, $N$ denotes the number of sampled predictions, whereas $k$ in pass@$k$ denotes the number of distinct ranked candidates.

For the HRM baseline and C2 checkpoints at step 4886, evaluation checks covered grid reconstruction, transformation inversion, candidate deduplication and ranking, and query- and task-level aggregation. Grid comparisons were normalised to a common integer dtype. An independent recomputation from saved predictions reproduced the counts obtained with this dtype-normalised evaluator. For both TRM runs, scores at $k = 1$ and $k = 2$ were independently recomputed from the saved submissions using integer-normalised grid comparisons, confirming 0/419 solved queries and 0/400 entirely solved tasks. For URM, augmentation identifiers were decoded to associate predictions with their original queries and invert the augmentation transformations. Input and target reconstruction were checked on known cases. Both 60-epoch checkpoints (step 4807) were evaluated using this procedure. For both 60-epoch URM configurations, ARC/pass@$k$ scores at $k = 1, 2, 100$ and $1000$ were separately recomputed from the saved predictions using integer-normalised grid comparisons and matched the evaluator outputs. The checks confirmed coverage of all 400 tasks and 419 queries, with all 385,814 augmented sequences associated with their original queries, and reproduced the first two candidates in the saved submissions. This score recomputation reused the dataset's geometric-inversion helper and therefore does not constitute an independent validation of every decoding component. For both 120-epoch URM checkpoints (step 9614), separate recomputation from saved predictions reproduced ARC/pass@1, @2, @100 and @1000, confirming coverage of all 400 tasks and 419 original queries, with no unassigned augmented sequences. The first two ranked candidates also matched the saved submissions. This recomputation reused the dataset's geometric-inversion helper and therefore does not independently validate every decoding component. Validated original-query scores remain unavailable for the damped configuration.

TRM evaluation loads the checkpoint at 50,000 optimisation steps. Across the official HRM, TRM and URM comparisons, baseline and nested configurations execute 16 recurrent segments per sequence, with adaptive early halting disabled during evaluation. Token accuracy and LM loss are computed from the final segment, rather than averaged across segments. Original-query candidates also derive from final-segment predictions. The evaluated URM checkpoints were saved through the EMA checkpointing path, with an EMA decay of 0.999. Checkpoints and evaluation procedures are associated with each reported result. Where multiple seeds are available, individual outcomes and observed variability are reported. Single-seed comparisons remain descriptive and exploratory. Augmented predictions from the same task are not treated as independent training replicates.\\

Follow-up experiments were defined in response to observations from the initial comparisons. They examined gradient propagation, state dynamics and sensitivity to the training budget.

In HRM, learning difficulties in C1 motivated the C1--C2 comparison, which extends the differentiable path while preserving forward computation. This comparison is available at seed 0. Runs of the baseline and C2 with seeds 0 and 1 additionally describe the observed variability in these configurations.

Two exploratory analyses are described in Appendix~B: a segment-wise audit of HRM trajectories, motivated by recent evidence that correct predictions can be lost across recursive segments \cite{ren_are_2026}, and URM diagnostics and interventions on input coupling, the number of core applications and intermediate-state damping.

Continued improvement towards the end of the 60-epoch URM regime motivated a baseline--variant comparison at 120 epochs, trained from scratch. This experiment examines sensitivity to the optimisation budget without assuming that insufficient training explains the initial outcomes. Follow-up findings are interpreted within their specific interventions, budgets and available repetitions.\\

\section{Results}

The main results compare the baseline and nested configurations within each model family using the official codebases under the adopted training budgets. The HRM C1--C2 comparison examines gradient propagation with unchanged forward computation, while the URM 60- and 120-epoch comparisons examine sensitivity to training duration. Exploratory results from adapted implementations are reported in Appendix~A, and trajectory audits and URM intervention diagnostics in Appendix~B.

\begin{table}[H]
\centering
\small
\setlength{\tabcolsep}{4pt}
\renewcommand{\arraystretch}{1.3}
\begin{tabular}{|p{0.22\textwidth}|p{0.07\textwidth}|p{0.28\textwidth}|p{0.20\textwidth}|p{0.13\textwidth}|}
\hline
\textbf{Configuration} &
\textbf{Seed} &
\textbf{Training budget} &
\textbf{Token accuracy (\%)} &
\textbf{LM loss} \\
\hline
HRM baseline & 0 & 61 epochs & 60.60 & 1.447 \\
\hline
HRM-F C1 & 0 & 61 epochs & 40.70 & 1.982 \\
\hline
HRM-F C2 & 0 & 61 epochs & 60.33 & 1.558 \\
\hline
HRM baseline & 1 & 61 epochs & 58.06 & 1.583 \\
\hline
HRM-F C2 & 1 & 61 epochs & 42.39 & 1.933 \\
\hline
TRM baseline & 0 & 50,000 optimisation steps & 59.86 & 1.420 \\
\hline
TRM-F & 0 & 50,000 optimisation steps & 58.59 & 1.491 \\
\hline
URM baseline & 0 & 60 epochs & 62.20 & 1.341 \\
\hline
URM-F & 0 & 60 epochs & 61.89 & 1.342 \\
\hline
URM baseline & 0 & 120 epochs & 62.79 & 1.295 \\
\hline
URM-F & 0 & 120 epochs & 62.80 & 1.286 \\
\hline
\end{tabular}
\vspace{8pt}

\caption{\textit{Official-codebase results under the adopted training budgets. Parameter counts were matched within each family. Core-application counts and gradient propagation are specified separately in the methodology. Metrics are compared within families; the table does not establish a ranking across architectures. Differences reported below are calculated before rounding. Token accuracy is averaged per sequence over valid labelled positions, including grid delimiters where present. LM loss denotes predictive stablemax cross-entropy, averaged over non-ignored target positions within each sequence and then across valid sequences, using outputs from the final recurrent segment of evaluation and excluding auxiliary halting and continuation losses.}}
\label{tab:official_results}
\end{table}

\begin{figure}[H]
    \centering
    \includegraphics[width=\textwidth]{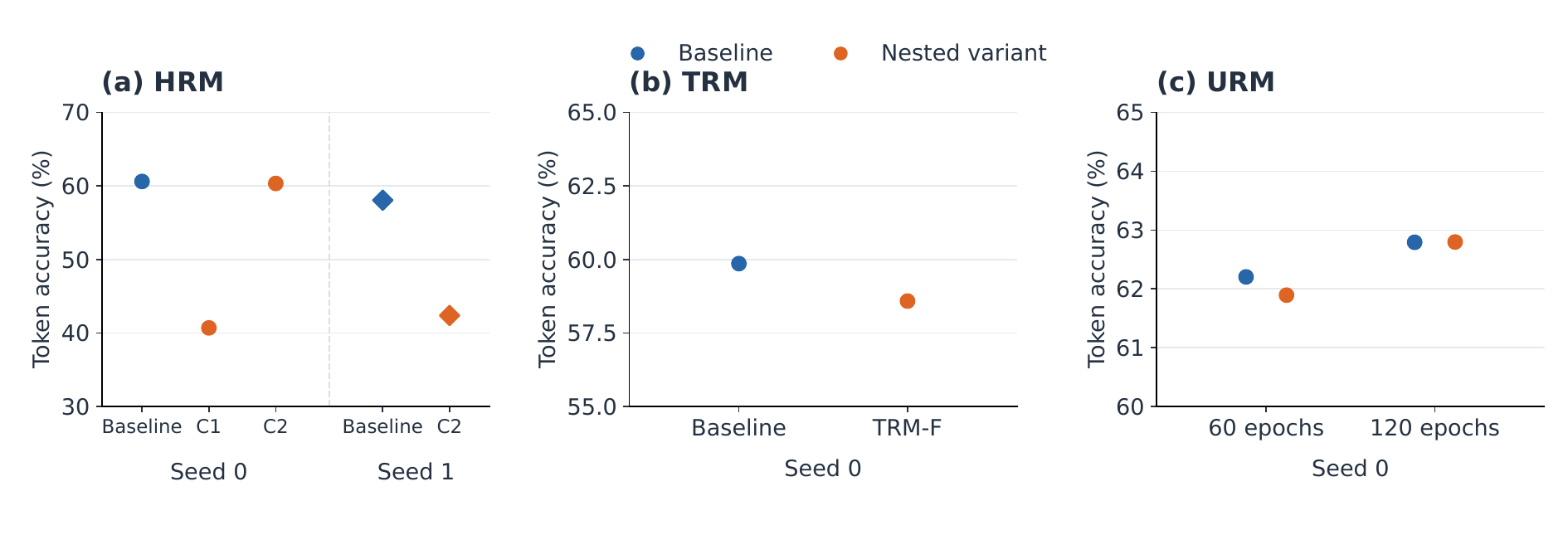}
    \caption{\textit{Token accuracy for the official-codebase comparisons on ARC-AGI-1 (Table~3). Blue denotes baseline configurations and orange denotes nested variants. \textbf{(a) HRM}: baseline, C1 and C2 at seed 0; baseline and C2 at seed 1. Circles denote seed 0 and diamonds denote seed 1; the dashed vertical line separates the seeds. C1 and C2 share the same forward-update schedule but differ in gradient propagation. \textbf{(b) TRM}: baseline and nested variant at seed 0. \textbf{(c) URM}: baseline and nested variant at 60 and 120 training epochs, seed 0. Vertical-axis ranges differ across panels. Token accuracy measures partial predictive correctness and is distinct from exact solving.}}
    \label{fig:figure2}
\end{figure}
\textbf{HRM.} C1 and C2 retained the same forward computation but differed in the differentiable segment of the nested recursion. At seed 0, extending gradient propagation in C2 was associated with an accuracy increase of 19.64 percentage points relative to C1, bringing it within 0.27 points of the baseline, although its loss remained higher. This proximity was not reproduced at seed 1, where C2 remained 15.67 points below the baseline. Original-task score at $k$ was reported for the baseline and C2 at seed 0: both obtained zero at $k = 1$ and $k = 2$. At $k = 10$, the baseline fully solved one task, with correct outputs for all its test queries among the first ten candidates; C2 did so only at $k = 1000$. At $k = 1000$, both configurations solved 2/419 original queries and 1/400 tasks completely. C1 and the seed-1 runs were compared using aggregate metrics.

\textbf{TRM.} At 50,000 optimisation steps, TRM-F achieved an accuracy 1.27 percentage points below the baseline, a higher loss and a lower exact accuracy on augmented sequences. Both configurations obtained zero ARC/pass@1 and ARC/pass@2. At $k = 100$ and $k = 1000$, the baseline reached 0.875\%, compared with 0.125\% for the variant.

\begin{table}[H]
\centering
\setlength{\tabcolsep}{4pt}
\renewcommand{\arraystretch}{1.3}
\resizebox{\textwidth}{!}{%
\begin{tabular}{|l|c|c|c|c|c|c|c|}
\hline
\textbf{Configuration} &
\shortstack{\textbf{Augmented-sequence}\\\textbf{exact accuracy (\%)}} &
\shortstack{\textbf{ARC/pass@1}\\\textbf{(\%)}} &
\shortstack{\textbf{ARC/pass@2}\\\textbf{(\%)}} &
\shortstack{\textbf{ARC/pass@5}\\\textbf{(\%)}} &
\shortstack{\textbf{ARC/pass@10}\\\textbf{(\%)}} &
\shortstack{\textbf{ARC/pass@100}\\\textbf{(\%)}} &
\shortstack{\textbf{ARC/pass@1000}\\\textbf{(\%)}} \\
\hline
TRM baseline & 0.01374 & 0 & 0 & 0.125 & 0.125 & 0.875 & 0.875 \\
\hline
TRM-F & 0.00052 & 0 & 0 & 0 & 0 & 0.125 & 0.125 \\
\hline
\end{tabular}%
}
\vspace{8pt}

\caption{\textit{Exact prediction metrics for the official TRM comparison at 50,000 training steps, seed 0. Augmented-sequence exact accuracy requires every valid labelled position to be correct. Augmented-sequence exact accuracy is given in percent of 385,815 evaluated sequences (53 and 2 exact sequences, respectively). ARC/pass@$k$ measures the mean per-task fraction of original queries solved by one of the first $k$ selected candidates; it does not represent the percentage of entirely solved tasks.}}
\label{tab:trm_exact}
\end{table}

\textbf{URM.} At 60 epochs, token accuracy was 62.20\% for the baseline and 61.89\% for URM-F, while predictive losses were close. Augmented-sequence exact accuracy was higher for URM-F than for the baseline (62 versus 10 correct sequences out of 385,814; Table~5). On original queries, URM-F correctly predicted one of 419 queries with its first candidate, yielding an ARC/pass@1 of 0.125\% under the per-task aggregation, whereas the baseline scored zero. This was the second query of a two-query task; neither configuration entirely solved a task at $k = 1$. The baseline ranked the same correct answer second, and both configurations reached 0.125\% at $k = 2$ and $k = 100$. At $k = 1000$, the baseline reached 0.750\%, with four correct queries and three entirely solved tasks, while URM-F remained at 0.125\%, with one correct query and no entirely solved tasks. Thus, the relative performance depended on $k$: URM-F placed the shared correct answer first, whereas the baseline recovered more correct answers within the larger candidate list. These single-seed results do not establish a reproducible advantage attributable to nesting.

Additional runs trained from scratch for 120 epochs examined performance under a larger training budget using seed 0. Accuracy increased from 62.20\% to 62.79\% for the baseline and from 61.89\% to 62.80\% for URM-F. At 120 epochs, the accuracy difference was less than 0.01 percentage points in favour of the variant, with LM losses of 1.295 and 1.286, respectively. Sequence exact accuracy increased for the baseline but decreased for the variant. At 120 epochs, both configurations scored zero at ARC/pass@1 and ARC/pass@2. The baseline reached 1.125\% at $k = 100$ and 1.250\% at $k = 1000$, compared with 0.625\% at both thresholds for URM-F. At $k = 1000$, these results corresponded to seven solved queries and five entirely solved tasks for the baseline, versus four queries and two entirely solved tasks for URM-F. Compared with 60 epochs, both configurations improved ARC/pass@1000, but URM-F's first-candidate success and the baseline's success at $k = 2$ were not retained.

Exploratory diagnostics are detailed in Appendix~B. The HRM trajectory audit did not show better preservation of exact predictions for C2 in the sampled training-pipeline sequences. URM interventions examined input coupling, additional core applications and intermediate-state damping. Damping improved diagnostic cross-entropy on training-pipeline batches but yielded worse full-evaluation accuracy and loss than unmodified URM-F. These assessments used different data and procedures. Original-query ARC/pass@$k$ scores are not reported for the damped variant.

\begin{table}[H]
\centering
\setlength{\tabcolsep}{4pt}
\renewcommand{\arraystretch}{1.3}
\resizebox{\textwidth}{!}{%
\begin{tabular}{|l|c|c|c|c|c|c|}
\hline
\textbf{Configuration} &
\shortstack{\textbf{Training}\\\textbf{budget}} &
\shortstack{\textbf{Augmented-sequence}\\\textbf{exact accuracy (\%)}} &
\shortstack{\textbf{ARC/pass@1}\\\textbf{(\%)}} &
\shortstack{\textbf{ARC/pass@2}\\\textbf{(\%)}} &
\shortstack{\textbf{ARC/pass@100}\\\textbf{(\%)}} &
\shortstack{\textbf{ARC/pass@1000}\\\textbf{(\%)}} \\
\hline
URM baseline & 60 epochs & 0.00259 & 0 & 0.125 & 0.125 & 0.750 \\
\hline
URM-F & 60 epochs & 0.01607 & 0.125 & 0.125 & 0.125 & 0.125 \\
\hline
URM baseline & 120 epochs & 0.01089 & 0 & 0 & 1.125 & 1.250 \\
\hline
URM-F & 120 epochs & 0.00855 & 0 & 0 & 0.625 & 0.625 \\
\hline
\end{tabular}%
}
\vspace{8pt}

\caption{\textit{Exact prediction metrics for the official-codebase URM comparisons, seed 0. Augmented-sequence exact accuracy requires every valid labelled position to be correct and is expressed as a percentage of 385,814 evaluated sequences. At 60 epochs, the baseline and URM-F correctly predict 10 and 62 augmented sequences, respectively. ARC/pass@$k$ is the mean per-task fraction of original queries solved by at least one of the first $k$ ranked candidates, expressed as a percentage; it is distinct from the percentage of entirely solved tasks. The 60- and 120-epoch results correspond to checkpoint steps 4807 and 9614, respectively.}}
\label{tab:urm_exact}
\end{table}

\section{Discussion}

The tested configurations showed no consistent predictive advantage over their within-family baselines on ARC-AGI-1 under the tested conditions. They document the behaviour of specific configurations and identify factors that remain coupled to the intervention.

The main limitation is the lack of informative first-candidate exact-solving performance. First-candidate exact solving was zero in the examined official-codebase HRM and TRM evaluations. At 60 epochs, URM-F solved one original query with its first candidate, whereas the URM baseline ranked the same correct answer second. The baseline recovered more correct queries within the first 1000 candidates. Thus, URM-F's advantage at $k = 1$ did not extend to larger candidate budgets. These sparse, single-seed URM outcomes neither establish equivalence nor demonstrate a reproducible benefit attributable to nesting. For HRM, original-query evaluations covered only the baseline and C2 at seed 0.

The architectural comparisons also leave several explanations unresolved. Parameter matching did not imply computational matching: official nested HRM used twelve lower-operator applications per $H$ cycle, compared with two in the baseline. Official TRM and URM matched core-application counts, but runtime, memory and total ACT computation were not equalised. All three variants also changed state coupling and direct input injection. In TRM, even with matched parameter counts, core applications and differentiable application counts, information flow differed between configurations. Its single-seed baseline advantage therefore describes the tested configurations without isolating the effect of nesting. These limitations motivate an HRM control without $M$ that retains the baseline's $L$--$H$ organisation and matches the variant's number of lower-operator applications, together with retrained comparisons that vary the intermediate-state organisation and input coupling separately.

The exploratory HRM gradient-path comparison showed a marked difference at seed 0, but its reproducibility across seeds has not been assessed. With forward computation unchanged, extending the differentiable segment from C1 to C2 was associated with an accuracy recovery of approximately twenty percentage points at seed 0. However, C1 was trained only at seed 0, and C2's proximity to the baseline did not recur at seed 1. The observation motivates replication of the gradient-path comparison, rather than a general claim that longer differentiable horizons improve nested models. A separate trajectory audit found no advantage of C2 in preserving exact predictions within the sampled training sequences. Learning through a longer differentiable segment and retaining correct predictions across ACT segments are distinct questions.

One possible interpretation is that nesting changes the learning problem as well as the forward update schedule. HRM motivates its one-step gradient approximation through an idealised fixed-point argument \cite{wang_hierarchical_2025}, whereas TRM questions those assumptions and reports benefits from differentiating all updates in the final recursive cycle \cite{jolicoeur-martineau_less_2025}. The C1--C2 observation is consistent with sensitivity to the differentiable horizon, but does not establish fixed-point convergence or explain the recovery mechanistically. Similarly, persistent states updated by a shared operator need not learn complementary functions. Because intermediate states receive input indirectly in the tested variants, state organisation and information flow remain alternative or interacting explanations. Both interpretations require controlled experiments.

The exploratory results illustrate additional limits on interpretation. In the final rapid HRM series (Appendix~A), cell-weighted accuracy and equal-task averaging gave different seed-level comparisons. URM damping improved diagnostic cross-entropy on training-pipeline batches but worsened full-evaluation loss, measured on different data. Because these assessments differed in data and procedures, the diagnostic improvement did not establish better original-query prediction. These findings motivate explicit aggregation rules and a predefined primary endpoint, with diagnostic measures retained as secondary evidence.

Budget, replication and development choices further restrict interpretation. Increasing URM training from 60 to 120 epochs improved token accuracy, predictive loss and ARC/pass@1000 in both configurations. However, both scored zero at ARC/pass@1 and ARC/pass@2 at 120 epochs, and augmented-sequence exact accuracy decreased for URM-F. These results show that improvements in average predictive metrics and candidate recovery need not translate into better first-candidate exact solving. This does not establish convergence or imply that further training would favour nesting. Official comparisons used one or two seeds, and the rapid comparisons used four or five usable repetitions. Nested HRM configurations showed greater dispersion across the available seeds than their baselines. Development-task reuse, training-pipeline diagnostics and selection of the damping coefficient on the diagnostic sample limit the independence of the exploratory evidence. Evaluation covered only ARC-AGI-1. Paired analyses can quantify task-level differences where predictions are available, but cannot recover uncertainty across training seeds when replicated runs are absent.

\section{Conclusions and Future Directions}

The tested nested configurations produced no consistent predictive advantage on ARC-AGI-1 at matched parameter counts. The available evidence is insufficient to support the architectural hypothesis, but limited exact solving, sparse replication and remaining confounding factors prevent a general conclusion against it. These findings are limited to the recursive-core configurations evaluated here.

The priority is a controlled within-family comparison testing whether adding a persistent intermediate state and organising lower-level recursive updates around it, using the same operator for the fast and intermediate states, improves exact solving relative to the family's baseline, at matched parameter counts and under comparable data and computational budgets. This question applies to HRM, TRM and URM, but the first confirmatory study should focus on one family. The choice should be made on development data, based on informative baseline exact solving and the feasibility of the required controls, before examining final evaluation results.

The first requirement is an evaluation pipeline checked against known outputs and, where available, a reference checkpoint with established performance. Development experiments should then identify a training regime in which baseline exact solving allows the planned tasks and training seeds to distinguish a predefined useful improvement from no effect. If this prerequisite cannot be met, the comparison remains inconclusive. The family and training regime should not be selected according to an observed advantage of the nested variant.

Within the selected family, baseline and nested configurations should use the same data and supervision, matched parameter counts, predefined comparable computational budgets and multiple training seeds. A separate final evaluation set should remain excluded from configuration selection. The design should include controls for the factors that accompany nesting in that family. If HRM is selected, a control without the intermediate state $M$ should retain the baseline's $L$--$H$ organisation while matching the variant's number of lower-operator applications. Across families, retrained controls should distinguish nesting from changes in state coupling and direct input injection. Gradient-path comparisons should preserve forward computation and be replicated across seeds. Runtime, peak memory and executed recurrent segments should accompany operator-application counts, since equal counts do not establish equal computational cost.

The primary metric will be the mean per-task fraction of original queries solved by the first selected candidate, with uniform candidate-selection and aggregation rules. Before the confirmatory experiments, the protocol should fix the minimum useful improvement in absolute percentage points, the replication plan and the uncertainty analysis. Analyses should respect grouping by task and distinguish variation across tasks from variation across training seeds \cite{agarwal_deep_2021,bouthillier_accounting_2021}. Pass@2 and partial token metrics may be reported as secondary outcomes. Configurations, controls and interpretation criteria should be specified before final evaluation.

Three outcomes would guide interpretation. A benefit reaching the minimum useful improvement, recurring across seeds and persisting under the controls would support the hypothesis in the studied regime. Estimates sufficiently precise to exclude that improvement would weaken it in that regime. An apparent advantage that disappears under a control would not support a nesting-specific benefit in that comparison. Insufficient baseline exact solving, or estimates too uncertain to distinguish a useful benefit from its absence, would leave the question unresolved. Additional resources could enable a more informative test, but would not guarantee confirmation.

Once an informative comparison has been completed in one family, extending the protocol to the others would assess whether its conclusions depend on the architecture. Further extensions should remain secondary: comparisons on ARC-AGI-2 \cite{chollet_arc-agi-2_2026}, Sudoku-Extreme and Maze-Hard \cite{wang_hierarchical_2025}; evaluation of fixed checkpoints with additional recurrent updates \cite{bansal_end--end_2022} and variation in the number of nested scales and the allocation of updates across them. These experiments would examine generality and sensitivity after the main comparison has established what can be concluded about nesting in the selected regime.

\section*{Code availability}

The code accompanying this study is available at \url{https://github.com/saraneri01/fractal-recursion/tree/main}. Documentation and usage instructions are provided in the repository README.










\bibliographystyle{unsrtnat}
\bibliography{references}

\appendix
\section*{Appendix A. Exploratory results from adapted implementations}
Preliminary development used adapted HRM and TRM implementations, termed rapid, with explicit demonstration conditioning. Checks covered parameter counts, forward and backward execution, evaluation masks, gradient norms and small-set memorisation. These implementations also supported exploratory optimisation experiments. The quantitative comparisons reported below comprise the final five-seed HRM series on 80 development tasks and the four-seed TRM series on 400 evaluation tasks. They remain distinct from the official-codebase comparisons because their model components, data representations and training procedures differ.

The examined adapted implementations use softmax cross-entropy averaged over the full output canvas, including padding, without explicit class weighting. Excluding padding from evaluation metrics therefore does not imply its exclusion from the training objective. In the deep-supervision scripts, each segment produces a separate optimiser update, after which the carried states are detached.

Content accuracy denotes the fraction of correctly predicted cells within the target output grid, excluding padding positions. Cell-weighted content accuracy pools cells across all evaluated tasks, so that larger grids carry more weight; the equal-task variant first averages accuracy within each task and then across tasks.

The examined raw data files contain 400 ARC-AGI-1 training tasks and 400 evaluation tasks, with no overlap in task identifiers. The reconstructed development split (seed 0, fold 0) selects 80 tasks from the training partition and has no task-identifier overlap with the 400 evaluation tasks.

Preliminary rapid HRM runs explored optimisation and supervision choices and are not reported quantitatively. The final series used a different training budget and evaluation set and therefore does not isolate the effects of individual development choices.

The final rapid HRM series used five seeds, a shared learning rate of $2.5 \times 10^{-5}$ and four supervised segments per batch, giving 40,000 optimiser updates over 10,000 outer steps. Both configurations used the same outer-update schedule ($N = 3$, $T = 4$), with three $H$ updates per forward call. Mean cell-weighted content accuracy was 72.79\% for the baseline and 72.30\% for the variant. The variant performed better in three seeds but showed a wider accuracy range across seeds (10.64 versus 5.75 percentage points). With equal weight assigned to each task, the baseline performed better in every seed. Task-level advantages recurred in at least three seeds for 28 tasks favouring the baseline and eight favouring the variant. Neither configuration solved any of the 80 development tasks exactly. These findings remain exploratory because this set also informed configuration choices.

Rapid TRM was evaluated over four seeds and 20,000 training steps per configuration. Mean content accuracy was 77.55\% for the baseline and 76.26\% for the variant, with the baseline ahead in every seed. The baseline solved one of the 400 evaluation tasks in one seed; the variant solved none. Configuration-specific learning rates, different core-application counts (27 in the variant, 21 in the baseline) and different differentiable segments (three versus seven applications) limit attribution of this gap to recursive organisation.

\section*{Appendix B. Exploratory trajectory and URM intervention diagnostics}

Recent mechanistic analysis of HRM reported that correct predictions can be lost during subsequent recursive segments \cite{ren_are_2026}. We therefore audited segment-wise trajectories of the official HRM baseline and C2 at seed 0, using the seed-0 \texttt{step\_4886} checkpoints from the 61-epoch official comparison. Both checkpoints were evaluated with fixed weights on the same 3,792 augmented sequences drawn from one sampled pass of the training loader, and each sequence was followed for 16 ACT segments. A sequence was considered exact at a given segment when all positions with valid labels were correctly predicted; this criterion concerns individual sequences and does not correspond to solving an original ARC task. For each sequence, we recorded whether it was exact at any segment, whether it was exact at the final segment, and whether an exact prediction subsequently became incorrect, either temporarily or through the final segment.

In the trajectory audit, the baseline produced an exact prediction at some segment for 29 of 3,792 sequences and at the final segment for 24. C2 reached 32 and 20, respectively. Among sequences that were exact at least once, correctness was lost at least temporarily in 10 of 29 baseline sequences (34.5\%) and 18 of 32 C2 sequences (56.3\%); the loss persisted to the final segment in 5 and 12 cases. In this sample, C2 therefore did not show better preservation of correct answers. The sets of sequences that became exact may differ between the two models, the counts are small, and the audit uses training-pipeline data from a single seed; these figures do not estimate general rates.

URM diagnostics characterised internal-state changes, magnitudes and representational similarity. For the input-injection, six-versus-eight-application and retrained damping comparisons reported below, diagnostic cross-entropy uses softmax over float32 logits. It is averaged over valid target tokens within each batch, then equally across batches. Target padding is excluded; non-ignored special tokens are included. Predictions are obtained after one recurrent segment from freshly initialised states, using small samples from the training pipeline. These values are not directly comparable with the full-evaluation LM loss, which uses stablemax, per-sequence normalisation and predictions after 16 recurrent segments on the augmented evaluation set.

Exploratory sensitivity tests examined whether modifying direct input injection, the number of core applications or intermediate-state update magnitude affected prediction quality. Fixed-weight interventions measured immediate responses to modified dynamics, while separately trained configurations assessed adaptation during learning. Damping modifies each intermediate-state update according to
\[
\tilde{z}_M = F(z_M + z_L), \quad
z_M \leftarrow (1 - \alpha)z_M + \alpha\tilde{z}_M
\]
where $F$ is the shared core and $\alpha$ is a fixed configuration value weighting its proposed state. Thus, $\alpha = 0.25$ retains 75\% of the previous state and incorporates 25\% of the proposal.

Beyond the HRM gradient-path comparison reported in Section~4, URM diagnostics motivated interventions on input coupling, update magnitude and computation. Two exploratory URM interventions examined input flow and the number of core applications. With checkpoint weights fixed, adding direct input to intermediate-state updates yielded a mean diagnostic cross-entropy of 13.48 on three training-pipeline batches of two sequences each, compared with 3.56 for unmodified URM-F evaluated on the same checkpoint with the same procedure; batch correspondence between the two runs was reconstructed rather than recorded. This comparison describes the response to modified dynamics without assessing adaptation through retraining. A separately trained pilot using eight shared-applications per group was compared with URM-F using six applications per group on the same 30 training-pipeline batches of two sequences each. With two groups per recurrent segment, these configurations perform 16 and 12 core applications per segment, respectively. The eight-application configuration had lower diagnostic cross-entropy in only 8/30 batches and a higher mean cross-entropy overall (3.963 versus 3.116). Neither investigation constituted a full evaluation of original queries, and their outcomes therefore remain diagnostic.

Damping the intermediate state produced a diagnostic benefit. Among four damping coefficients tested at fixed weights, a value of 0.25 yielded the lowest mean diagnostic cross-entropy. After retraining, the damped variant also achieved lower diagnostic cross-entropy in 20 of 30 two-sequence batches at 30 epochs and 23 of 30 at 60 epochs. These counts describe differences on training-pipeline batches, not independent training repetitions. This diagnostic benefit was not accompanied by higher token accuracy or lower LM loss on the full evaluation set: at 60 epochs, the damped variant achieved 61.26\% token accuracy and an LM loss of 1.655, compared with 61.89\% and 1.342 for original URM-F. Its augmented-sequence exact accuracy was 0.00674\%.

\end{document}